\documentclass[10pt,twocolumn,letterpaper]{article}

\usepackage{cvpr}              

\usepackage{multirow}
\usepackage{comment}
\newcommand{\midnet}{waypoint network\xspace}
\newcommand{\midneta}{WayNet\xspace}
\newcommand{\latentmap}{memory proxy map\xspace}
\newcommand{\latentmapa}{MPM\xspace}
\definecolor{cvprblue}{rgb}{0.21,0.49,0.74}
\usepackage[pagebackref,breaklinks,colorlinks,allcolors=cvprblue]{hyperref}

\def\paperID{14488} 
\def\confName{CVPR}
\def\confYear{2026}

\title{FeudalNav: A Simple Framework for Visual Navigation}

\author{Faith Johnson\\
Rutgers University\\
{\tt\small faith.johnson@rutgers.edu}
\and
Bryan Bo Cao\\
Stony Brook University\\
{\tt\small boccao@cs.stonybrook.edu}
\and
Shubham Jain\\
Stony Brook University\\
{\tt\small jain@cs.stonybrook.edu}
\and
Ashwin Ashok\\
Georgia State University\\
{\tt\small aashok@gsu.edu}
\and
Kristin Dana\\
Rutgers University\\
{\tt\small kristin.dana@rutgers.edu}
}

\begin{document}
\maketitle
\begin{abstract}

Visual navigation for robotics is inspired by the human ability to navigate environments using visual cues and memory, eliminating the need for detailed maps. In unseen, unmapped, or GPS-denied settings, traditional metric map-based methods fall short, prompting a shift toward learning-based approaches with minimal exploration.
In this work, we develop a hierarchical framework\footnote{Code is available at \href{https://github.com/visnavdev/feudalnav}{https://github.com/visnavdev/feudalnav}.} that decomposes the navigation decision-making process into multiple levels. Our method learns to select subgoals through a simple, transferable waypoint selection network. A key component of the approach is a latent-space  memory module  organized solely by visual similarity, as a proxy for distance.
This alternative to  graph-based topological representations proves sufficient for navigation tasks, 
providing a compact, light-weight, simple-to-train navigator that can find its way to the goal in novel locations. We show competitive results with a suite of SOTA methods in Habitat AI environments without using any odometry in training or inference.  An additional contribution leverages the interpretability of the framework for interactive navigation. We consider the question:  \textit{how much direction intervention/interaction is needed to achieve success in all trials?}  We demonstrate that even minimal human involvement can significantly enhance overall navigation performance. 
\end{abstract}    
\section{Introduction}
\label{sec:intro}

Visual navigation is based on the idea that humans navigate without constructing detailed 3D maps of their environment. Instead, we explore, wander, and form approximate mental maps to aid navigation. In psychology, the concept of cognitive maps and graphs \cite{tolman1948cognitive, chrastil2014cognitive, peer2021structuring, epstein2017cognitive} formalizes this intuition, with studies showing that humans encode relative distances between landmarks as approximate graphs.

In vision and robotics, these ideas have been translated into topological graphs and latent maps derived from visual observations. These {\it visual navigation} paradigms aim to develop representations that are semantically rich, easily updated, and more compact to store than full 3D metric maps \cite{Gupta_2017_CVPR, savinov2018semi, chaplot2020neural, mirowski2018learning, Savarese-RSS-19, gervet2023navigating, he2023metric, kim2023topological}.

Reinforcement Learning (RL) is commonly used for navigation in known environments, with distance-to-goal serving as a reward proxy in many RL-based navigation frameworks \cite{shah2021ving, eysenbach2019search}. In this work, we focus on visual navigation in environments without odometry data, presenting a more challenging approach than those assuming access to noise-free GPS+compass sensors, such as image-goal navigation challenges \cite{habitatchallenge2023}.

Many visual navigation methods rely on odometry, including SLAM-based approaches requiring known camera poses \cite{kwon2023renderable, chaplot2019learning}, graph-based methods using distance to define edge features \cite{kim2023topological}, and RL techniques with distance-based rewards \cite{wijmans2019dd}. Inspired by NRNS \cite{hahn2021no}, which questioned the necessity of RL for exploration in visual navigation, our method eliminates both. We take this further by removing odometry and graphs entirely, leveraging a hierarchical framework, waypoint prediction map, and memory proxy map in 2D latent space. We demonstrate that a high-performing visual navigation agent can be built without odometry, RL, graphs, or metric maps.

\begin{figure}
    \centering
   \includegraphics[width=\linewidth]{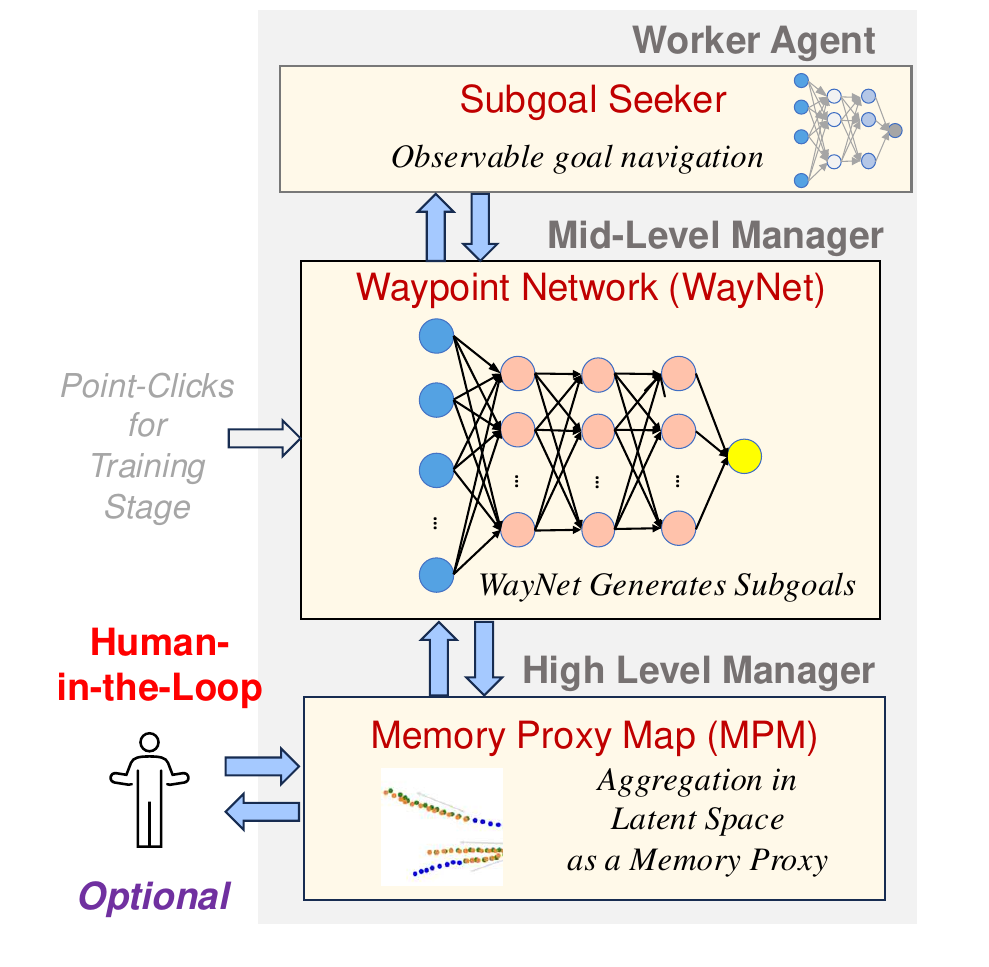}
   \vspace{-14pt}
    \caption{
     FeudalNav provides a no-graph, no-odometry, and no-RL visual navigation agent for the image-goal task on previously unseen environments. This simple framework uses a  hierarchy that consists of: (1) a high-level manager with a memory proxy map (MPM) that frames memory as a latent space learning problem, (2) a mid-level manager \midnet (\midneta) mimicking human teleoperation to guide worker agent exploration, and (3) a low-level worker choosing actions in the environment based on the previous layers' subgoals. Optionally, a human-in-the-loop component intervene to improve navigation (see Section~\ref{sec:navhf}). 
   }
   \label{fig:overview}
    \vspace{-15pt}
\end{figure}

Our hierarchical structure is inspired by feudal learning \cite{vezhnevets2017feudal, dayan1992feudal}, which decomposes tasks into sub-components, offering performance advantages well suited for visual navigation shown in Figure \ref{fig:overview}. The feudal framework defines {\it workers} and {\it managers} across multiple hierarchical levels (e.g., mid-level and high-level managers), with each entity observing different aspects of the task and operating at distinct temporal or spatial scales.
For navigation in unseen environments, this structure is particularly effective. The worker-agent handles local motion, while manager-agents oversee navigation and determine when to explore new regions, making the overall task more manageable.

A key aspect of our approach is representing the traversed environment with a learned latent map instead of a graph, serving as a sufficient memory proxy during navigation. This {\it \latentmap (\latentmapa)} is obtained through self-supervised contrastive learning. The high-level manager of our feudal learning agent maintains the \latentmapa, using MPM density to assess exploration levels and decide when to move to a new region.
Another core component is the {\it \midnet (\midneta)}, the mid-level manager that generates waypoints—visible sub-goals guiding the worker agent. We train \midneta via supervised learning to imitate human exploration policies using point-click navigation trajectories from the LAVN dataset \cite{johnson2024landmark}. When humans navigate via point-and-click teleoperation, they intuitively select a target point—such as the end of a hallway, a door, or deeper into a room. This subgoal selection skill is easily learnable and generalizes well to new environments with zero-shot transfer.
The mid-level manager selects subgoals, while a low-level worker chooses actions that avoid obstacles and navigates toward nearby observable subgoals. This streamlined framework achieves competitive SOTA performance on the image-goal navigation task in unseen Gibson environments \cite{xiazamirhe2018gibsonenv} within Habitat AI \cite{habitat19iccv}, a simulation environment built from real-world scans.


Our \textbf{contributions} are fourfold: \textbf{1)}  A self-supervised \latentmap (\latentmapa) that enables lean, no-odometry, no-graph, no-RL navigation, \textbf{2)} A \midnet (\midneta) for local navigation through supervised learning of human exploration policies, \textbf{3)} A hierarchical navigation framework using agents operating at different spatial scales, and \textbf{4)} SOTA performance on the image-goal task in different Habitat indoor environments.


\section{Related Work}
\textbf{Visual Navigation}
Visual navigation aims to build representations that incorporate the rich information of scenes by injecting image-based learning into traditional mapping and planning  navigation frameworks \cite{Gupta_2017_CVPR, chaplot2019learning, devo2020towards, shah2021ving,seymour2021maast}. 
Early work focused on creating full metric maps of a space using SLAM augmented by images \cite{chaplot2019learning,chaplot2020neural}. 
While full metric maps can be ideal, especially if the space can be mapped before planning, the representations are computationally complex.
Topological graphs and maps can lighten this load  and provide image data at nodes and relative distances at edges \cite{savinov2018semi,chen2019behavioral}.
While easier to build, these methods require odometry to be readily available and have the potential for large memory requirements, especially if new nodes are added every time an agent takes an action \cite{shah2021ving,shah2022offline,he2023metric}. 
One solution to this problem is sparser topological graphs \cite{hahn2021no,shah2021rapid} 
where visual features of unexplored next-nodes are sometimes predicted or hallucinated  \cite{he2023metric, bar2024navigation}.
Some methods go a step further by pairing semantic labels with the graph representation \cite{kim2023topological,chang2023goat}. However, these methods can break down in environments that are sparse or featureless, have many duplicate objects, or contain uncommon objects that may not appear in object detection datasets. Another solution is to only use graphs during training to build 2D embedding space representations of environments, i.e.\ potential fields or functions, that preserve important physical \cite{morin2023one, ramakrishnan2022poni,bono2023learning}, visual \cite{henriques2018mapnet,bono2023end,ramakrishnan2022environment}, or semantic \cite{georgakis2021learning,chaplot2020object,al2022zero} relationships between regions in the environment.  
\begin{figure}[h]
    \centering
    \includegraphics[width=\linewidth]{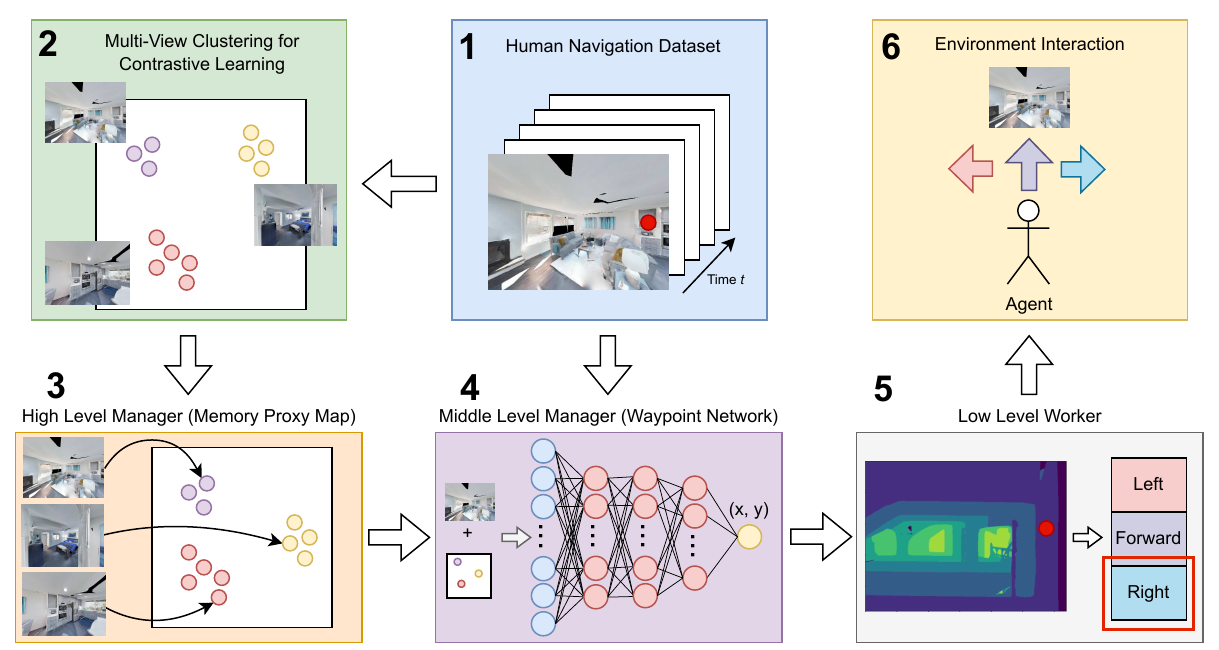}
   \caption{Method Overview \textbf{1:} A subset of trajectories of point-click and observation-image pairs are selected from the LAVN dataset \cite{johnson2024landmark} 
   for learning a latent space for the memory proxy map and training WayNet. We test our method on a separate set of environments. \textbf{2:} Images from these pairs are clustered based on feature similarity, and cluster members form positive pairs used for contrastively learning a latent space. \textbf{3:} The learned latent space is used to build a memory proxy map where the high level manager (HLM) records a history of agent locations. 
   \textbf{4:} The waypoint network (Waynet) is trained 
   to provide subgoals (points) for navigation based on visual observations, imitating human teleoperation via point-clicks. \textbf{5:} Based on this point-click guidance and depth map input, the low-level worker predicts 
   to either more forward, left, or right
   in order to move towards the subgoal (point) and avoid obstacles. \textbf{6:} During test time, these low level actions guide agent movement and produce new observations as input for the upper levels of the hierarchy. }
    \label{fig:processdiagram}
\end{figure}
Our method more closely aligns with this line of work,
but we do not use any graph networks or graph inference and instead
 build our 2D latent map using self-supervised contrastive methods. 
Additionally, unlike many of the methods above, we do not require the agent to have information about the test environment (odometry) before deployment, do not use RL or graphs, or learn 3D metric maps. 
\vspace{-0.1in}
\paragraph{Feudal Learning}
Feudal learning originated as a reinforcement learning (RL) framework \cite{dayan1992feudal,vezhnevets2017feudal}. Researchers have explored RL for the image-goal visual navigation task \cite{zhu2017target}, most notably using external memory buffers \cite{kumar2018visual, fang2019scene, beeching2020egomap, mezghan2022memory}.
However, it still suffers from several issues such as sample inefficiency, handling sparse rewards, and the long horizon problem \cite{fujimoto2021minimalist,le2018hierarchical}. 
Feudal reinforcement learning, characterized by its composition of multiple, sequentially stacked agents working in parallel, arose to combat these issues using temporal or spatial abstraction, mostly in simulated environments \cite{vezhnevets2017feudal} where its effects can be more easily compared to other methods. Some of these task hierarchies are hand defined \cite{vezhnevets2020options}, while others are discovered dynamically with \cite{chen2020ask} or without \cite{li2020hrl4in} human input. 
The feudal network paradigm has been adopted by other learning schemas outside of RL in recent years, as the hierarchical network structure provides benefits to other methodologies outside of RL. In navigation, hierarchical networks are commonly used to propose waypoints as subgoals during navigation \cite{chane2021goal}, typically working in a top-down view of the environment \cite{xu2021hierarchical} with only two levels of agents \cite{wohlke2021hierarchies}. Our work uses multiple agent levels, operates in the first person point of view for predicting waypoints, and reaps the benefits of this feudal relationship without using reinforcement learning as shown in Figure \ref{fig:processdiagram}. 

\section{Methods}
\label{sec:method}
\begin{figure}[h]
    \centering
    \vspace{-15pt}
    \includegraphics[width=0.91\linewidth]{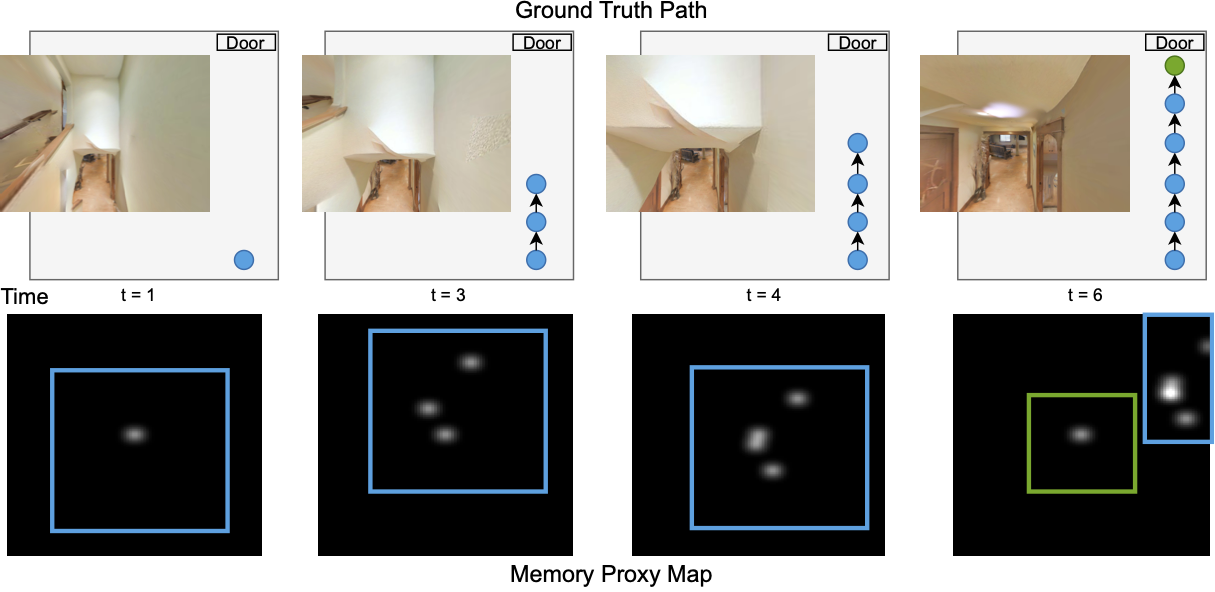}
    \caption{Illustration of the memory proxy map (MPM) during navigation.
Row 1: RGB observation images along a trajectory are shown alongside a diagram of the agent's location in the environment. Colored circles (blue/green) represent the traveled path.
Row 2: The MPM with Gaussian-weighted occupancy markers corresponding to each observation image. The map is local, fixed in size, and cropped around the most recent latent map position. This allows the agent to mark locations in latent space (rather than metric space) and recognize repeated observations.
Similar observation images cluster together (blue) until a significantly different view appears, forming a new group (green). The MPM provides a graph-free (and interpretable) mechanism for tracking previously visited regions, proving efficient in image-goal navigation by quantifying exploration in different areas of the environment.}
    \vspace{-10pt}
    \label{fig:makingMPM}
\end{figure}

\paragraph{\bf High-Level Manager: Memory}
We contrastively learn a latent space to construct an aggregate {\it \latentmap (\latentmapa)} demonstrated in Figure \ref{fig:makingMPM}, serving as a memory module for our feudal navigation agent. This self-supervised latent space is trained using Synchronous Momentum Grouping (SMoG) \cite{pang2022smog}, which combines instance-level contrastive learning with clustering. SMoG’s momentum grouping enables simultaneous instance- and group-level contrastive learning, and we empirically demonstrate its effectiveness in construcing the learned latent space comprising the MPM.
Instead of traditional data augmentations (e.g., rotation, scaling, shifting) to define positive pairs for contrastive learning, we dynamically form clusters of visually similar images observed along training trajectories. These clusters are determined using SuperGlue \cite{sarlin2020superglue} for robust keypoint matching and are used to define positive and negative pairs during training. This approach  allows the latent space to preserve a proxy for  relative distance between images without requiring ground truth odometry.

For each training trajectory, the first observed image initializes a cluster center. As the agent moves, each new image is compared to a memory bank of cluster centers. SuperGlue obtains keypoint matches and if the confidence $\alpha_c$ is high, the image is added to an existing cluster; otherwise, it forms a new cluster center. Clusters are built per environment across all training trajectories, and images from the same cluster are randomly sampled as positive pairs for contrastive training.
During inference, observation images are dynamically embedded in the contrastively learned latent space to construct a {\it \latentmap} (MPM) of previously visited locations as the agent navigates novel environments. To enhance interpretability, we introduce an {\it isomap imitator network} for a 2D projection. We train an MLP to map the learned SMoG feature space (128 dimensions) to a 2D representation.
Specifically, we compute isomap 2D embeddings for all training data to reduce dimensionality while preserving relative feature distances. Since isomap produces different coordinates with each run, a simple MLP is trained to replicate the isomap embeddings from SMoG features.
To update the {\it \latentmapa} during inference, a Gaussian-weighted circular window with $\sigma=1$ is applied to the predicted 2D location from the isomap imitator for each observation, generating a density map that reflects the level of exploration in different areas.

As an area is explored, the {\it \latentmapa} becomes more densely populated with Gaussian occupancy markers, helping the agent localize itself relative to past observations while also quantifying exploration levels.
This approach enables efficient navigation, serving as a robust and effective memory proxy.
\vspace{-0.2in}
\paragraph{\bf Mid-Level Manager: Direction}
For the mid-level manager, we leverage human knowledge to learn optimal navigation and exploration policies from demonstrations in the human navigation dataset. The intuition is that point-click navigation decisions in LAVN \cite{johnson2024landmark} are learnable and generalize to new environments with zero-shot transfer.
For each LAVN trajectory, we use the HLM to generate a memory proxy map. Using RGBD observations and corresponding {\it \latentmapa} crops centered on the agent’s location, we fine-tune ResNet-18 \cite{he2016deep} to predict a pixel coordinate guiding the agent’s motion. To make the feudal agent goal-directed, which is essential for image-goal navigation, we compute keypoint matches between the current observation and the goal image using SuperGlue \cite{sarlin2020superglue}. If the confidence $\alpha_k$ is high, the average of the matched keypoints replaces the mid-level manager's waypoint prediction.
This setup allows the agent to mimic human navigation while actively verifying if it has reached the goal. The high-level manager operates at a global, coarse-grained scale, while the mid-level manager (WayNet) has a first-person, fine-grained perspective. This hierarchical division introduces spatial abstraction, enabling the two networks to work in tandem, solving smaller subproblems efficiently. As a result, this approach outperforms end-to-end implementations  while allowing for faster training with less data (see Section~\ref{sec:results}).
\vspace{-0.2in}
\paragraph{\bf Low-Level Worker: Action}
The low level worker agent takes actions in the environment based on the current depth map and the waypoint predicted by WayNet. We define the following action space in accordance with other SOTA methods \cite{hahn2021no, yadav2022offline, wasserman2023last}: ``turn left $15^\circ$'', ``turn right $15^\circ$'', and ``move forward $0.25m$''.  Although an RL agent is typically used for this task, we find an MLP classifier works well to enable effective navigation. 
This classifier is trained to learn a mapping between the human-chosen action from the LAVN dataset \cite{johnson2024landmark} and the corresponding depth map and waypoint input. 
The agent stops navigation when the confidence threshold $\alpha_m$ for matching goal image features to the current observation is high and either the agent's depth measurement indicates it is sufficiently close to the goal location ($\leq1m$) or the ratio $\psi$ of the area of the matched keypoints to the total image size  is relatively large.
\section{Results}
\label{sec:results}
\paragraph{Image-Goal Navigation Task}
We test the performance of our method (FeudalNav) using the procedure outlined in NRNS \cite{hahn2021no} on the image-goal task in previously unseen Gibson \cite{xiazamirhe2018gibsonenv} test environments. To start, the agent is placed in an environment and given RGBD image observations of the first person view of their surroundings and a goal location. All images are $480 \times 640$ pixels with $120^\circ$ field of view. A trial terminates if the agent stops within $1m$ of the location of the goal image or the agent takes $500$ actions in the environment.
Each agent trajectory is evaluated on success rate (whether or not the goal has been reached) and SPL (success weighted by inverse path length), defined as 
\vspace{-3pt}
\begin{equation}
 SPL = \frac{1}{N}\sum^N_{i=0}S_i\frac{l_i}{\max(l_i,p_i)}
\end{equation}
where $N$ is the total number of trajectories considered, $S_i$ is an indicator variable for success, $l_i$ is the optimal (shortest) geodesic path length between the starting location and the goal, and $p_i$ is the actual path length the agent traveled.  \vspace{-0.1in}
\paragraph{Training and Testing Procedure}

We train our method with the LAVN \cite{johnson2024landmark} human navigation dataset, using a subset of 117 trajectories with a total of 36834 frames. The average number of contrastive learning training data clusters per trajectory is 23 (median 25). We train all models on either a GTX TITAN X or RTX 2080 Ti using a learning rate of $1$e-$4$. The confidence thresholds $\alpha_c$, $\alpha_k$, and $\alpha_m$ are chosen empirically to be $0.7$, and $\psi$ is similarly chosen as $0.85$. The isomap imitator network used to make the MPM is a two layer MLP with ReLU activations trained for 2000 epochs with batch size$~=32$, and we train SMoG \cite{pang2022smog} for 50 epochs with batch size$~=16$. We use the current observation and a $480 \times 640$ pixel crop of the MPM centered around the agent's current observation as input to WayNet, a modified ResNet-18 \cite{he2016deep} that accepts 7 channel input trained for 250 epochs with batch size$~=16$. The classifier network for the low-level worker is a four layer MLP with PReLU activations where the depth map and waypoint input have distinct projection heads trained for 2 epochs with batch size$~=128$. In total, our method trains in 3M iterations. Compared to other SOTA that uses RL and simulators and trains for anywhere from 10-100M iterations \cite{wijmans2019dd,al2022zero,hahn2021no} (or even 50 GPU days \cite{yadav2022offline}) using anywhere from 14.5-500M frames on up to 64 GPUs, our network consumes several orders of magnitude less data, significantly less GPUs, and a fraction of the total iterations.

We test our network using the testing procedure and baselines outlined in NRNS \cite{hahn2021no}. 
Testing trajectories come from a publicly available set of Gibson \cite{xiazamirhe2018gibsonenv} environments listed in \cite{hahn2021no}. They consist of approximately $6K$ point pairs (start and goal locations) that are uniformly sampled from fourteen environments and divided between two curvatures (straight/curved) and three goal distances ($1.5-3m$, $3-5m$, $5-10m$).  
We compare our method performance against flat RL DD-PPO \cite{wijmans2019dd} trained for varying lengths of time , behavior cloning (BC) with a ResNet-18 backbone and either a GRU or a metric map from \cite{hahn2021no}, NRNS \cite{hahn2021no} with and without noise, ZSEL \cite{al2022zero}, OVRL \cite{yadav2022offline}, and NRNS and OVRL enhaced by SLING \cite{wasserman2023last}. 
These methods are either trained directly in a simulator or for extended periods of time (ie. 100M time steps, 53 GPU days), require odometry, or use graphs in their implementations. 

There are other recent works that test on the image-goal task that require testing in previously seen environments, full scene reconstruction \cite{kwon2023renderable}, or full panoramic or semantic images \cite{kim2023topological} that are very reliant of the fixed spacing or semantic contextual information of residential dwellings. These are excluded as  unfair comparisons 
since our method does not assume prior knowledge of semantic context (since it is not readily available in many applications).
\begin{figure}[h]
    \centering
    \includegraphics[width=\linewidth]{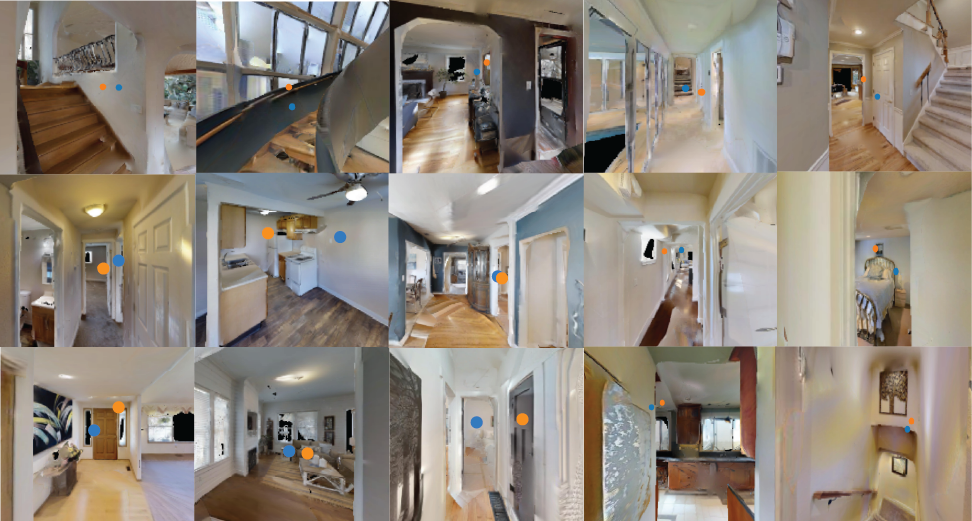}
    \vspace{-10pt}
    \caption{(Best viewed zoomed) We show qualitative results for the waypoints predicted by \midneta (blue) shown with the ground truth human click points from the LAVN dataset \cite{johnson2024landmark} (orange). Note that the majority of the samples show high overlap between the two. When they diverge, the \midneta waypoints still lead to navigably feasible areas in each observation, showing that our network sufficiently learns an acceptable navigation policy.}
    \vspace{-10pt}
    \label{fig:mlmPoints}
\end{figure}
Figure \ref{fig:mlmPoints} shows qualitative examples of the mid-level manager's predicted waypoints (blue) shown with the human ground truth point clicks (orange) from the LAVN dataset \cite{johnson2024landmark}. For the majority of the points, both the predictions and the ground truth lie near the same area or on the same object, or the prediction identifies another feasibly navigable area. 

\begin{table*}[t]
    \centering
    \small
    \setlength{\tabcolsep}{4.3pt}
    \renewcommand{\arraystretch}{1.05}
    \begin{tabular}{l|p{16em}|cc|cc|cc||cc}
    \toprule
    \multirow{2}{*}{} & \multirow{2}{*}{\textbf{Model}} & 
    \multicolumn{2}{c|}{\textbf{Easy}} & 
    \multicolumn{2}{c|}{\textbf{Medium}} & 
    \multicolumn{2}{c||}{\textbf{Hard}} & 
    \multicolumn{2}{c}{\textbf{Average}} \\ 
    \cmidrule(lr){3-10}
     & & Succ$\uparrow$ & SPL$\uparrow$ & Succ$\uparrow$ & SPL$\uparrow$ & Succ$\uparrow$ & SPL$\uparrow$ & Succ$\uparrow$ & SPL$\uparrow$ \\
    \midrule
    \midrule
    \multirow{8}{*}{\rotatebox[origin=c]{90}{\textbf{Straight}}} 
     & DDPPO (10M steps)* & 10.5 & 6.7 & 18.1 & 16.2 & 11.8 & 10.9 & 13.5 & 11.2 \\
     & DDPPO (extra data + 50M steps)* & 36.3 & 34.9 & 35.7 & 34.0 & 6.0 & 6.3 & 26.0 & 25.1 \\
     & DDPPO (extra data + 100M steps)* & 43.2 & 38.5 & 36.4 & 35.0 & 7.4 & 7.2 & 29.0 & 26.9 \\
     & BC w/ ResNet + Metric Map & 24.8 & 24.0 & 11.5 & 11.3 & 1.4 & 1.3 & 12.6 & 12.2 \\
     & BC w/ ResNet + GRU & 34.9 & 33.4 & 17.6 & 17.1 & 6.1 & 5.9 & 19.5 & 18.8 \\
     & NRNS w/ noise & 64.1 & 55.4 & 47.9 & 39.5 & 25.2 & 18.1 & 45.7 & 37.7 \\
     & NRNS w/out noise & 68.0 & 61.6 & 49.1 & 44.6 & 23.8 & 18.3 & 47.0 & 41.5 \\
     & NRNS + SLING & \textbf{85.3} & 74.4 & 66.8 & 49.3 & 41.1 & 28.8 & 64.4 & 50.8 \\
     & OVRL + SLING* & 71.2 & 54.1 & 60.3 & 44.4 & 43.0 & 29.1 & 58.2 & 42.5 \\
     & \textbf{FeudalNav (Ours)} & 82.6 & \textbf{75.0} & \textbf{71.0} & \textbf{57.4} & \textbf{49.0} & \textbf{34.2} & \textbf{67.5} & \textbf{55.5} \\
    \midrule
    \multirow{10}{*}{\rotatebox[origin=c]{90}{\textbf{Curved}}}
     & DDPPO (10M steps)* & 7.9 & 3.3 & 9.5 & 7.1 & 5.5 & 4.7 & 7.6 & 5.0 \\
     & DDPPO (extra data + 50M steps)* & 18.1 & 15.4 & 16.3 & 14.5 & 2.6 & 2.2 & 12.3 & 10.7 \\
     & DDPPO (extra data + 100M steps)* & 22.2 & 16.5 & 20.7 & 18.5 & 4.2 & 3.7 & 15.7 & 12.9 \\
     & BC w/ ResNet + Metric Map & 3.1 & 2.5 & 0.8 & 0.7 & 0.2 & 0.2 & 1.4 & 1.1 \\
     & BC w/ ResNet + GRU & 3.6 & 2.9 & 1.1 & 0.9 & 0.5 & 0.4 & 1.7 & 1.4 \\
     & NRNS w/ noise & 27.3 & 10.6 & 23.1 & 10.4 & 10.5 & 5.6 & 20.3 & 8.8 \\
     & NRNS w/out noise & 35.5 & 18.4 & 23.9 & 12.1 & 12.5 & 6.8 & 24.0 & 12.4 \\
     & ZSEL* & 41.0 & 28.2 & 27.3 & 18.6 & 9.3 & 6.0 & 25.9 & 17.6 \\
     & OVRL* (53 GPU days) & 53.6 & 31.7 & 47.6 & 30.2 & 35.6 & 21.9 & 45.6 & 28.0 \\
     & NRNS + SLING & 58.6 & 16.1 & 47.6 & 16.8 & 24.9 & 10.1 & 43.7 & 14.3 \\
     & OVRL + SLING* \cite{wasserman2023last} & 68.4 & 47.0 & 57.7 & 39.8 & 40.2 & \textbf{25.5} & 55.4 & 37.4 \\
     & \textbf{FeudalNav (Ours)} & \textbf{72.5} & \textbf{51.3} & \textbf{64.4} & \textbf{40.7} & \textbf{43.7} & 25.3 & \textbf{60.2} & \textbf{39.1} \\
    \bottomrule
    \end{tabular}
    \caption{\textbf{Quantitative results on the image-goal navigation task.} 
    Following NRNS~\cite{hahn2021no}, we evaluate in unseen Gibson environments~\cite{xiazamirhe2018gibsonenv}. 
    We compare our \textbf{FeudalNav} with baselines (DDPPO~\cite{wijmans2019dd}, NRNS~\cite{hahn2021no}) and SOTA methods (ZSEL~\cite{al2022zero}, OVRL~\cite{yadav2022offline}, SLING~\cite{wasserman2023last}). 
    Bold numbers denote the best results. 
    (* indicates simulator-based training).}
    \label{tab:NRNSTable}
    \vspace{-6pt}
\end{table*}

\vspace{-0.1in}
\paragraph{FeudalNav Performance}
We show performance on the image-goal task as detailed in section 4 for our feudal navigation agent (FeudalNav) and SOTA in Table \ref{tab:NRNSTable}. We also report performance averaged across the easy, medium, and hard trials, which we use to conduct comparisons of model performance. Our method has shown a significant improvement in success rate performance of $108\%$ (straight) and $283\%$ (curved) over all DDPO \cite{wijmans2019dd} baselines and  $208\%$ (straight) and $3380\%$ (curved) over both behavior cloning (BC) methods \cite{hahn2021no} while using no RL, learning no metric map, and not training directly in a simulator. 
We achieve a $5\%$ increase in performance on average success rate over NRNS+SLING \cite{wasserman2023last} and a $16\%$ increase over OVRL+SLING \cite{wasserman2023last} on straight trajectories. We make larger improvements to SPL performance as well ($9\%$ and $31\%$ over NRNS+SLING and OVRL+SLING respectively), despite not using odometry, not explicitly parsing semantics, not utilizing a graph, and only using $\sim37 K$ images for training (compared to the 3.5 million used by NRNS and 14.5 million used by OVRL).
This trend of performance improvement continues for the curved case where FeudalNav has a $9\%$ increase on average success rate over NRNS+SLING and $ 38\%$ over OVRL+SLING. This corresponds with a $173\%$ and $5\%$ increase in SPL against these two methods respectively.  
In real applications, it is less likely that a robot will be tasked to find an object within a straight line of sight from itself. For this reason, performance on the curved trajectory case gives a more realistic prediction of real world performance. 

\begin{figure*}[]
    \centering
    \includegraphics[width=0.9\linewidth]{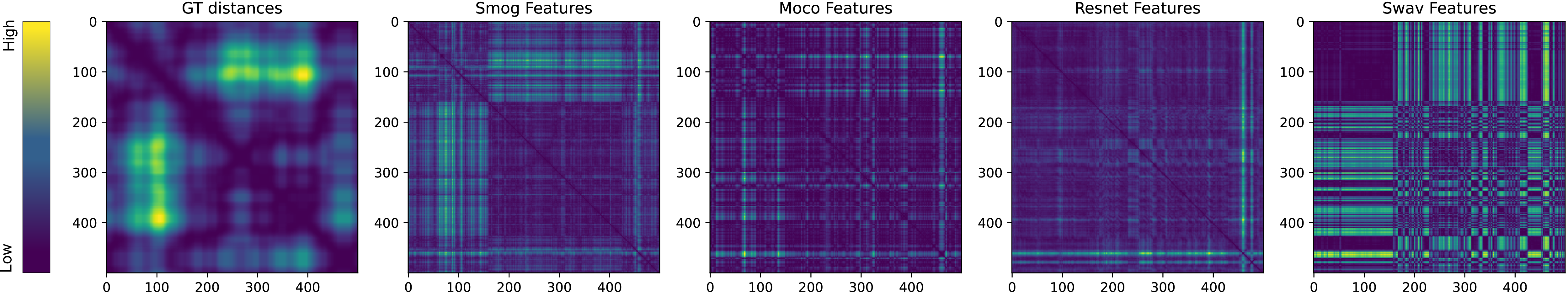}
    \caption{
    Heatmaps of distance matrices illustrate metric distances between image pairs in a 450-image trajectory. Left: The ground truth distance matrix (brighter indicates greater distance) represents spatial separation in the simulated environment. Right: Feature-distance matrices (computed via MSE) compare image pairs using features from SmOG \cite{pang2022smog}, MoCo \cite{he2020momentum}, ResNet \cite{he2016deep}, and SwAV \cite{caron2020unsupervised}.
 We select SMoG for feature detection as good proxy for metric distance.
    }
    \label{fig:hlmFeats}
\end{figure*}

\vspace{-0.1in}

\paragraph{Feature Comparison}
 We postulate that image feature distances can be indicative of  ground truth distances between points. In Figure \ref{fig:hlmFeats}, we compare the ground truth distances between observation pairs from a single trajectory in the Copemish Gibson environment of the LAVN dataset \cite{johnson2024landmark} to the feature distances predicted by our high-level manager, which utilizes SMoG \cite{pang2022smog}. For baselines, we also evaluate inter-observation feature distance matrices for MoCo \cite{he2020momentum}, ResNet-18 \cite{he2016deep}, and SwAV \cite{caron2020unsupervised}.
In the figure, lower distances appear darker/purple, while higher distances are lighter/yellow. A latent space that closely mirrors geometric distances between images should resemble the first square, which represents ground truth distances. SwAV \cite{caron2020unsupervised} learns highly diverse features, leading to large feature distances, whereas ResNet \cite{he2016deep} has the opposite issue—highly similar features across the trajectory—due to the relatively low diversity in indoor scene views compared to its training data. Both fail to serve as effective distance proxies for navigation.
MoCo \cite{he2020momentum} begins to approximate the ground truth distance matrix but still produces overly similar features. Validating our intuition, SMoG \cite{pang2022smog} best preserves relative distances, as its optimization of inter-sample and inter-cluster distances enables a latent space well-suited for this purpose.

\begin{table*}[t]
    \centering
    \footnotesize
    \setlength{\tabcolsep}{4.0pt}
    \renewcommand{\arraystretch}{1.05}
    \begin{tabular}{l|c|c|c|cc|cc|cc||cc}
    \toprule
    \multirow{2}{*}{} & \multirow{2}{*}{\textbf{\latentmapa}} & \multirow{2}{*}{\textbf{\midneta}} & \multirow{2}{*}{\textbf{LLW}} &
    \multicolumn{2}{c|}{\textbf{Easy}} &
    \multicolumn{2}{c|}{\textbf{Medium}} &
    \multicolumn{2}{c||}{\textbf{Hard}} &
    \multicolumn{2}{c}{\textbf{Average}} \\ 
    \cmidrule(lr){5-12}
    & & & & Succ$\uparrow$ & SPL$\uparrow$ & Succ$\uparrow$ & SPL$\uparrow$ & Succ$\uparrow$ & SPL$\uparrow$ & Succ$\uparrow$ & SPL$\uparrow$ \\
    \midrule
    \midrule
    \multirow{7}{*}{\rotatebox[origin=c]{90}{\textbf{Straight}}} 
     & -- & RGB & Det & 48.00 & 30.28 & 37.00 & 21.75 & 24.57 & 13.21 & 36.52 & 21.75 \\
     & -- & RGBD & Det & 48.20 & 31.70 & 39.40 & 21.50 & 24.94 & 12.99 & 37.51 & 22.06 \\
     & -- & 3 RGBD & Det & 50.20 & 31.19 & 38.60 & 21.24 & 20.72 & 9.16 & 36.51 & 20.53 \\
     & \textit{+} & RGBD & Det & 72.00 & 64.55 & 60.40 & 50.96 & 41.32 & 34.01 & 57.91 & 49.84 \\
     & \textit{+} & 3 RGBD & Det & 77.50 & 64.53 & 62.50 & 42.18 & 44.29 & 24.50 & 61.43 & 43.74 \\
     & \textbf{\textit{+}} & \textbf{RGBD} & \textbf{Cl} & \textbf{82.60} & \textbf{74.95} & \textbf{71.00} & \textbf{57.40} & \textbf{49.01} & \textbf{34.20} & \textbf{67.54} & \textbf{55.52} \\
     & \textit{+} & 3 RGBD & Cl & 73.60 & 73.05 & 37.10 & 35.66 & 9.18 & 8.95 & 39.96 & 39.22 \\
    \midrule
    \multirow{7}{*}{\rotatebox[origin=c]{90}{\textbf{Curved}}} 
     & -- & RGB & Det & 34.70 & 11.20 & 32.60 & 13.02 & 18.20 & 7.24 & 28.50 & 10.49 \\
     & -- & RGBD & Det & 36.60 & 11.55 & 30.00 & 11.86 & 18.30 & 7.72 & 28.30 & 10.37 \\
     & -- & 3 RGBD & Det & 39.50 & 11.91 & 32.80 & 11.75 & 15.70 & 5.65 & 29.33 & 9.77 \\
     & \textit{+} & RGBD & Det & 53.80 & 27.91 & 42.60 & 25.00 & 27.20 & 17.01 & 41.20 & 23.31 \\
     & \textit{+} & 3 RGBD & Det & 68.60 & 28.93 & 56.40 & 26.06 & 32.40 & 14.44 & 52.47 & 23.14 \\
     & \textbf{\textit{+}} & \textbf{RGBD} & \textbf{Cl} & \textbf{72.50} & \textbf{51.26} & \textbf{64.40} & \textbf{40.73} & \textbf{43.70} & \textbf{25.32} & \textbf{60.20} & \textbf{39.11} \\
     & \textit{+} & 3 RGBD & Cl & 59.00 & 55.52 & 15.50 & 14.58 & 1.50 & 1.45 & 25.33 & 23.85 \\
    \bottomrule
    \end{tabular}
    \caption{\textbf{Ablation study of FeudalNav components.} 
    We analyze the contribution of each module on the image-goal navigation task. 
    In the second column, ``--'' indicates models without the \textbf{MPM} (high-level manager), while ``+'' denotes the use of the \textbf{MPM} described in Section~3. 
    For \midneta, we use RGB or RGBD inputs, either from a single frame or from 3-frame history. 
    For the low-level worker (LLW), ``Det'' denotes a deterministic waypoint-to-action mapping, while ``Cl'' indicates the classification network in Section~3. 
    \textbf{The official FeudalNav configuration is shown in bold.} 
    Including \latentmap\ improves performance by providing observation frequency cues that guide attention to unexplored areas.}
    \label{tab:ablationTable}
    \vspace{-6pt}
\end{table*}

%
\vspace{-5pt}
\paragraph{Role of Full Hierarchy}
We conduct an ablation study to inspect the importance of each level of hierarchy in Table \ref{tab:ablationTable}. We incrementally add each piece of our architecture together and report the image-goal task results for the same experiment listed in Section 4. The second column indicates whether or not the high level manager's \latentmap (MPM) is included in the feudal navigation agent. Networks without this module are denoted by --  and \textit{+} denotes the MPM is used. For the mid-level manager, we test versions of WayNet that take in a single RGB, RGBD input and versions that take in three historical timesteps worth of each of these respective inputs (3 RGBD and 3 RGBD-M). We test two versions of the low-level worker: one that deterministically maps waypoint image coordinate predictions to environment actions (``Det") and one that follows the classifier approach detailed in Section \ref{sec:method} (``Cls").

For the networks without the MPM, we found that \midneta using RGB input performed qualitatively worse than using RGBD input, despite their similar performance, because depth information allowed the agent to avoid obstacles more efficiently. There is a $25\%$ increase in performance when the MPM is added to our network. This shows that the memory component is crucial to image-goal navigation success.  Furthermore, we find that a learning-based approach performs best for obstacle avoidance with a $10\%$ increase from the deterministic (``Det'') to the classification (``Cls'') low-level worker.
\begin{figure*}
    \centering
    \includegraphics[width=0.94\linewidth]{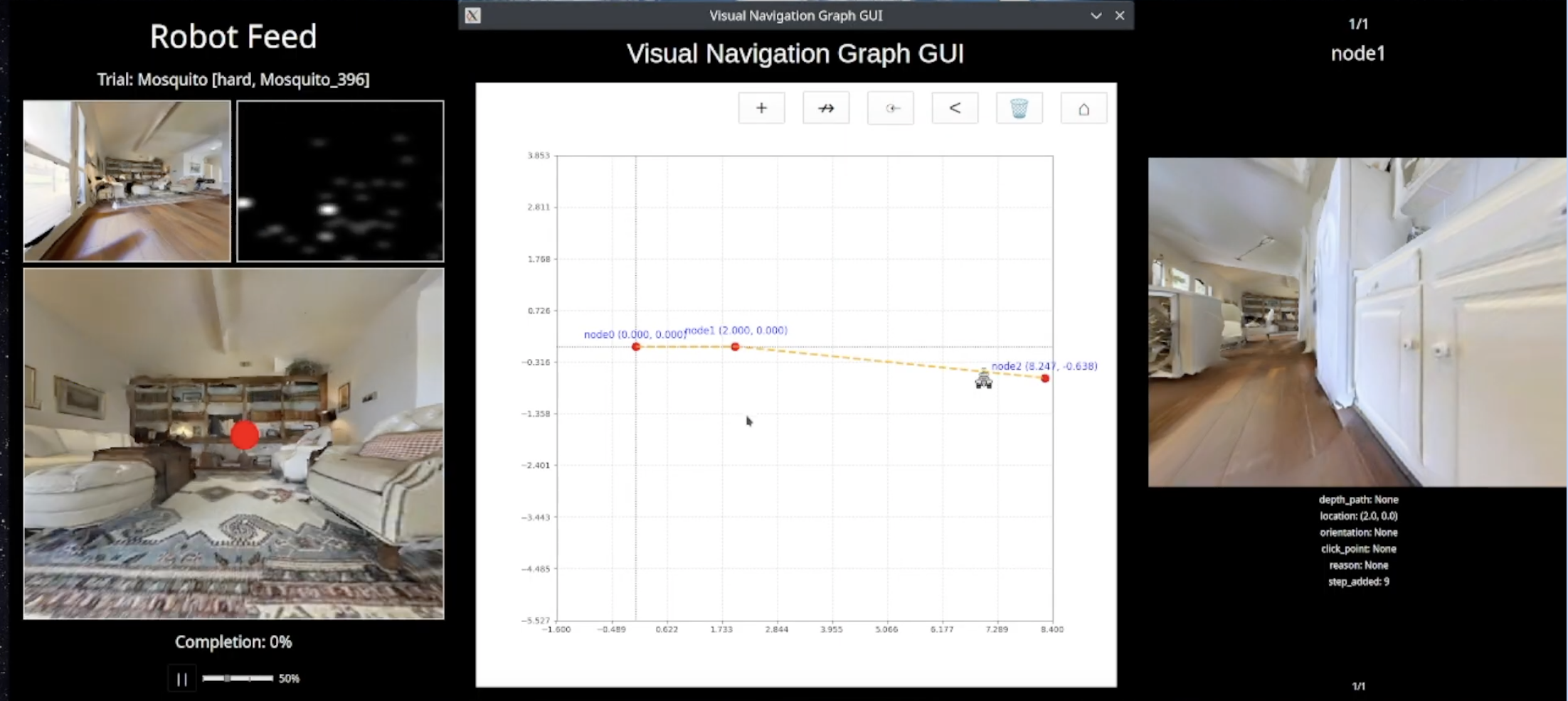}
    \caption{Interaction interface for the human partner/collaborator during the image-goal navigation task. The human interacts using the landmark graph (center) to assist the agent in finding the goal (top-left image). Clicking on a node displays the representative image (right) along with other details about the node. The MPM (to the right of the goal image) and the current robot video feed (bottom left) are also displayed. The red dot on the video feed indicates the waypoint predicted by Waynet.}
    \label{fig:hfView}
\end{figure*}
\subsection{Navigation with Human Feedback}
\label{sec:navhf}
While fully automated navigation is a common goal for intelligent agents, {\it human-robot partnered navigation} remains an unsolved challenge, crucial for applications like mobility assistance, agriculture, infrastructure inspection, and emergency response. We propose a human-in-the-loop framework that enables efficient visual navigation while simultaneously curating an interpretable {\it landmark graph} as an approximate mental map shared by both human and agent. This hierarchical approach allows collaboration, improving navigation success. When the agent makes a navigation mistake, the human intervenes to correct it, ensuring near-complete success with minimal intervention rather than tele-operation. Designed for convenience, the framework uses the landmark graph as a shared, actionable communication tool between human and agent.

\begin{table*}[t]
    \centering
    \small
    \setlength{\tabcolsep}{5pt}
    \renewcommand{\arraystretch}{1.05}
    \begin{tabular}{l|cc|cc|cc||cc}
    \toprule
    \textbf{Model} & 
    \multicolumn{2}{c|}{\textbf{Easy}} &
    \multicolumn{2}{c|}{\textbf{Medium}} &
    \multicolumn{2}{c||}{\textbf{Hard}} &
    \multicolumn{2}{c}{\textbf{Average}} \\
    \cmidrule(lr){2-9}
     & Succ$\uparrow$ & SPL$\uparrow$ & Succ$\uparrow$ & SPL$\uparrow$ & Succ$\uparrow$ & SPL$\uparrow$ & Succ$\uparrow$ & SPL$\uparrow$ \\
    \midrule
    ANS~\cite{chaplot2020learning} & 74.20 & 20.50 & 68.40 & 22.90 & 29.90 & 11.00 & 57.50 & 18.10 \\
    SPTM~\cite{savinov2018semi} & 66.50 & 40.60 & 64.20 & 38.50 & 42.10 & 25.40 & 57.60 & 34.80 \\
    Neural Planner~\cite{beeching2020learning} & 71.70 & 41.30 & 64.70 & 38.50 & 42.00 & 27.00 & 59.50 & 35.60 \\
    NTS~\cite{chaplot2020neural} & 87.00 & 65.00 & 58.00 & 38.00 & 43.00 & 26.00 & 63.00 & 43.00 \\
    MARL~\cite{mezghan2022memory} & 78.00 & 63.00 & 70.00 & 57.00 & 60.00 & 48.00 & 69.00 & 56.00 \\
    MemoNav~\cite{li2024memonav} & -- & -- & -- & -- & -- & -- & 74.70 & 57.90 \\
    VGM~\cite{kwon2021visual} & 86.00 & \textbf{80.00} & 81.00 & \textbf{68.00} & 61.00 & \textbf{46.00} & 76.10 & \textbf{64.50} \\
    \midrule
    \textbf{FeudalNav (Ours)} & 84.09 & 67.08 & 71.56 & 51.30 & 47.97 & 30.15 & 67.87 & 49.51 \\
    \textbf{FeudalNav + HF (Ours)} & \textbf{92.18} & 71.06 & \textbf{85.23} & 58.98 & \textbf{64.94} & 38.54 & \textbf{80.78} & 56.19 \\
    \bottomrule
    \end{tabular}
    \caption{\textbf{Comparison on the Gibson dataset test split for the image-goal navigation task, with and without human feedback (HF).}
    Results are reported in terms of success rate (Succ) and SPL. 
    Bold numbers indicate the best performance in each column.}
    \label{tab:HFGibsonTable}
    \vspace{-6pt}
\end{table*}

To construct the landmark graph, we first convert information from the high-level agent's memory proxy map (MPM). Landmarks are created by identifying clusters in the MPM based on visual similarity. Simultaneously, we leverage the noise-free actions of the Habitat AI \cite{habitat19iccv} simulator to approximate the agent's location. The start location is set to $(0,0)$, and the robot updates its estimated location after each action. When a new cluster is detected in the MPM, a corresponding node is added to the landmark graph at the estimated robot location. This landmark graph is then displayed to a human navigation partner alongside the current video feed, MPM, and goal image shown in Figure \ref{fig:hfView}.

The human updates the landmark graph by selecting whether navigation should move toward or away from a particular landmark. These instructions are relayed to the mid-level agent by directly modifying the MPM. If movement toward a landmark is indicated, points corresponding to its visual features are removed from the MPM. Conversely, if movement away is desired, the MPM is filled in. This allows the human to control exploration level in a given area.

Notably, communication is bidirectional: the human interprets and edits the landmark graph to reflect the scene and provide navigation hints, while the high-level agent adds landmarks when it detects distinct semantic structures, sharing its understanding of the environment. Thus, the continuously updated landmark graph serves as a {\it shared representation} between the human and the agent. 

To evaluate the human-robot dyad, we first run FeudalNav on the test set of the Gibson dataset \cite{xiazamirhe2018gibsonenv}, following other SOTA approaches \cite{chaplot2020learning,chaplot2020neural,li2024memonav,kwon2021visual}. We then rerun the failure cases, allowing the human to interact with FeudalNav as described above. This approach assumes the robot can solve a substantial portion of test cases independently, minimizing human intervention. By focusing on failure cases, we reduce the need for human input while gaining insights into the system's deficiencies.

We report the success rate and SPL for FeudalNav, the human-robot dyad (FeudalNav + HF), and other SOTA networks in Table \ref{tab:HFGibsonTable}, along with the average number of human interventions per trial is 6.11,4.33 and 3.91 for the easy, medium, and hard trials, respectively. 
FeudalNav alone outperforms ANS \cite{chaplot2020learning}, SPTM \cite{savinov2018semi}, Neural Planner \cite{beeching2020learning}, and NTS \cite{chaplot2020neural}. With human supervision (FeudalNav + HF), both success rate and SPL improve consistently, demonstrating that the human-robot dyad outperforms other SOTA methods. As expected, SPL gains are smaller than those in success rate because human interventions modify the MPM, altering the agent's exploration behavior. For instance, removing points from the MPM increases exploration time, leading to longer trajectories during trials.
\section{Conclusion}
Our work extends prior research on visual navigation for the image goal task by providing a high performance, no-RL, no-graph, no-odometry solution. The resulting methodology is efficient, lightweight and demonstrably effective. Our frameworks accomplishes this by putting emphasis on building representations for {\it agent memory}, i.e.\  the memory proxy map (MPM).  Such an emphasis is critically important for no-graph solutions and can be extended for new paradigms of visual navigation that include continual learning. 

\section*{Acknowledgments}This work was supported by the National Science Foundation (NSF) under grant NSF NRT-FW-HTF: Socially Cognizant Robotics for a Technology Enhanced Society (SOCRATES) No. 2021628 and grant nos. CNS-2055520, CNS-1901355, CNS-1901133.

\newpage~
{
    \small
    \bibliographystyle{ieeenat_fullname}
    \bibliography{bibfile}

@String(CVPR= {IEEE Conf. Comput. Vis. Pattern Recog.})

@String(ICCV= {Int. Conf. Comput. Vis.})

@String(ICLR = {Int. Conf. Learn. Represent.})

@String(CVPR  = {CVPR})

@String(ICCV  = {ICCV})

@String(ICLR  = {ICLR})

@inproceedings{kwon2021visual,
  title={Visual graph memory with unsupervised representation for visual navigation},
  author={Kwon, Obin and Kim, Nuri and Choi, Yunho and Yoo, Hwiyeon and Park, Jeongho and Oh, Songhwai},
  booktitle={Proceedings of the IEEE/CVF international conference on computer vision},
  pages={15890--15899},
  year={2021}
}

@article{johnson2024landmark,
  title={\href{https://arxiv.org/pdf/2402.14281}{A Landmark-Aware Visual Navigation Dataset}},
  author={Johnson, Faith and Cao, Bryan Bo and Dana, Kristin and Jain, Shubham and Ashok, Ashwin},
  journal={ACM Conference on Human Robot Interaction HRI 2025},
  year={2025}
}

@misc{habitatchallenge2023,
  title = {\href{https://aihabitat.org/challenge/2023/}{Habitat Challenge 2023}},
  author = {Karmesh Yadav and Jacob Krantz and Ram Ramrakhya and Santhosh Kumar Ramakrishnan and Jimmy Yang and Austin Wang and John Turner and Aaron Gokaslan and Vincent-Pierre Berges and Roozbeh Mootaghi and Oleksandr Maksymets and Angel X Chang and Manolis Savva and Alexander Clegg and Devendra Singh Chaplot and Dhruv Batra},
  howpublished = {\url{https://aihabitat.org/challenge/2023/}},
  year = {2023}
}

@article{chang2023goat,
  title={\href{https://arxiv.org/pdf/2311.06430}{Goat: Go to any thing}},
  author={Chang, Matthew and Gervet, Theophile and Khanna, Mukul and Yenamandra, Sriram and Shah, Dhruv and Min, So Yeon and Shah, Kavit and Paxton, Chris and Gupta, Saurabh and Batra, Dhruv and others},
  journal={arXiv preprint arXiv:2311.06430},
  year={2023}
}

@article{morin2023one,
  title={\href{https://arxiv.org/pdf/2303.04011.pdf}{One-4-All: Neural Potential Fields for Embodied Navigation}},
  author={Morin, Sacha and Saavedra-Ruiz, Miguel and Paull, Liam},
  journal={arXiv preprint arXiv:2303.04011},
  year={2023}
}

@inproceedings{ramakrishnan2022poni,
  title={\href{https://openaccess.thecvf.com/content/CVPR2022/papers/Ramakrishnan_PONI_Potential_Functions_for_ObjectGoal_Navigation_With_Interaction-Free_Learning_CVPR_2022_paper.pdf}{Poni: Potential functions for objectgoal navigation with interaction-free learning}},
  author={Ramakrishnan, Santhosh Kumar and Chaplot, Devendra Singh and Al-Halah, Ziad and Malik, Jitendra and Grauman, Kristen},
  booktitle={Proceedings of the IEEE/CVF Conference on Computer Vision and Pattern Recognition},
  pages={18890--18900},
  year={2022}
}

@article{hahn2021no,
  title={\href{https://proceedings.neurips.cc/paper_files/paper/2021/file/e02a35b1563d0db53486ec068ebab80f-Paper.pdf}{No rl, no simulation: Learning to navigate without navigating}},
  author={Hahn, Meera and Chaplot, Devendra Singh and Tulsiani, Shubham and Mukadam, Mustafa and Rehg, James M and Gupta, Abhinav},
  journal={Advances in Neural Information Processing Systems},
  volume={34},
  pages={26661--26673},
  year={2021}
}

@article{he2023metric, 
  title={\href{https://arxiv.org/pdf/2303.09192.pdf}{Metric-Free Exploration for Topological Mapping by Task and Motion Imitation in Feature Space}},
  author={He, Yuhang and Fang, Irving and Li, Yiming and Shah, Rushi Bhavesh and Feng, Chen},
  journal={arXiv preprint arXiv:2303.09192},
  year={2023}
}

@inproceedings{kim2023topological,
  title={\href{https://proceedings.mlr.press/v205/kim23a/kim23a.pdf}{Topological Semantic Graph Memory for Image-Goal Navigation}},
  author={Kim, Nuri and Kwon, Obin and Yoo, Hwiyeon and Choi, Yunho and Park, Jeongho and Oh, Songhwai},
  booktitle={Conference on Robot Learning},
  pages={393--402},
  year={2023},
  organization={PMLR}
}

@article{georgakis2021learning,
  title={\href{https://arxiv.org/pdf/2106.15648.pdf}{Learning to map for active semantic goal navigation}},
  author={Georgakis, Georgios and Bucher, Bernadette and Schmeckpeper, Karl and Singh, Siddharth and Daniilidis, Kostas},
  journal={arXiv preprint arXiv:2106.15648},
  year={2021}
}

@article{savinov2018semi,
  title={\href{https://arxiv.org/pdf/1803.00653.pdf}{Semi-parametric topological memory for navigation}},
  author={Savinov, Nikolay and Dosovitskiy, Alexey and Koltun, Vladlen},
  journal={arXiv preprint arXiv:1803.00653},
  year={2018}
}

@inproceedings{shah2021ving,
  title={\href{https://ieeexplore.ieee.org/stamp/stamp.jsp?arnumber=9561936}{Ving: Learning open-world navigation with visual goals}},
  author={Shah, Dhruv and Eysenbach, Benjamin and Kahn, Gregory and Rhinehart, Nicholas and Levine, Sergey},
  booktitle={2021 IEEE International Conference on Robotics and Automation (ICRA)},
  pages={13215--13222},
  year={2021},
  organization={IEEE}
}

@article{shah2022offline,
  title={\href{https://arxiv.org/pdf/2212.08244.pdf}{Offline reinforcement learning for visual navigation}},
  author={Shah, Dhruv and Bhorkar, Arjun and Leen, Hrish and Kostrikov, Ilya and Rhinehart, Nick and Levine, Sergey},
  journal={arXiv preprint arXiv:2212.08244},
  year={2022}
}

@article{shah2021rapid,
  title={\href{https://arxiv.org/pdf/2104.05859.pdf}{Rapid exploration for open-world navigation with latent goal models}},
  author={Shah, Dhruv and Eysenbach, Benjamin and Rhinehart, Nicholas and Levine, Sergey},
  journal={arXiv preprint arXiv:2104.05859},
  year={2021}
}

@inproceedings{henriques2018mapnet,
  title={\href{https://openaccess.thecvf.com/content_cvpr_2018/papers/Henriques_MapNet_An_Allocentric_CVPR_2018_paper.pdf}{Mapnet: An allocentric spatial memory for mapping environments}},
  author={Henriques, Joao F and Vedaldi, Andrea},
  booktitle={proceedings of the IEEE Conference on Computer Vision and Pattern Recognition},
  pages={8476--8484},
  year={2018}
}

@article{chrastil2014cognitive,
  title={\href{https://journals.plos.org/plosone/article/file?id=10.1371/journal.pone.0112544&type=printable}{From cognitive maps to cognitive graphs}},
  author={Chrastil, Elizabeth R and Warren, William H},
  journal={PloS one},
  volume={9},
  number={11},
  pages={e112544},
  year={2014},
  publisher={Public Library of Science San Francisco, USA}
}

@inproceedings{chaplot2020learning,
  title={LEARNING TO EXPLORE USING ACTIVE NEURAL SLAM},
  author={Chaplot, Devendra Singh and Gandhi, Dhiraj and Gupta, Saurabh and Gupta, Abhinav and Salakhutdinov, Ruslan},
  booktitle={8th International Conference on Learning Representations, ICLR 2020},
  year={2020}
}

@inproceedings{chaplot2019learning,
  title={\href{https://arxiv.org/pdf/2004.05155.pdf}{Learning To Explore Using Active Neural SLAM}},
  author={Chaplot, Devendra Singh and Gandhi, Dhiraj and Gupta, Saurabh and Gupta, Abhinav and Salakhutdinov, Ruslan},
  booktitle={International Conference on Learning Representations},
  year={2019}
}

@inproceedings{chaplot2020neural,
  title={\href{https://openaccess.thecvf.com/content_CVPR_2020/papers/Chaplot_Neural_Topological_SLAM_for_Visual_Navigation_CVPR_2020_paper.pdf}{Neural topological slam for visual navigation}},
  author={Chaplot, Devendra Singh and Salakhutdinov, Ruslan and Gupta, Abhinav and Gupta, Saurabh},
  booktitle={Proceedings of the IEEE/CVF Conference on Computer Vision and Pattern Recognition},
  pages={12875--12884},
  year={2020}
}

@article{eysenbach2019search,
  title={\href{https://proceedings.neurips.cc/paper/2019/file/5c48ff18e0a47baaf81d8b8ea51eec92-Paper.pdf}{Search on the replay buffer: Bridging planning and reinforcement learning}},
  author={Eysenbach, Ben and Salakhutdinov, Russ R and Levine, Sergey},
  journal={Advances in Neural Information Processing Systems},
  volume={32},
  year={2019}
}

@article{xu2021hierarchical,
  title={\href{https://arxiv.org/pdf/2106.03665.pdf}{Hierarchical robot navigation in novel environments using rough 2-d maps}},
  author={Xu, Chengguang and Amato, Christopher and Wong, Lawson LS},
  journal={arXiv preprint arXiv:2106.03665},
  year={2021}
}

@inproceedings{wohlke2021hierarchies,
  title={\href{https://ieeexplore.ieee.org/stamp/stamp.jsp?arnumber=9561151}{Hierarchies of planning and reinforcement learning for robot navigation}},
  author={W{\"o}hlke, Jan and Schmitt, Felix and van Hoof, Herke},
  booktitle={2021 IEEE International Conference on Robotics and Automation (ICRA)},
  pages={10682--10688},
  year={2021},
  organization={IEEE}
}

@article{fujimoto2021minimalist,
  title={\href{https://proceedings.neurips.cc/paper_files/paper/2021/file/a8166da05c5a094f7dc03724b41886e5-Paper.pdf}{A minimalist approach to offline reinforcement learning}},
  author={Fujimoto, Scott and Gu, Shixiang Shane},
  journal={Advances in neural information processing systems},
  volume={34},
  pages={20132--20145},
  year={2021}
}

@inproceedings{sarlin2020superglue,
  title={\href{https://openaccess.thecvf.com/content_CVPR_2020/papers/Sarlin_SuperGlue_Learning_Feature_Matching_With_Graph_Neural_Networks_CVPR_2020_paper.pdf}{Superglue: Learning feature matching with graph neural networks}},
  author={Sarlin, Paul-Edouard and DeTone, Daniel and Malisiewicz, Tomasz and Rabinovich, Andrew},
  booktitle={Proceedings of the IEEE/CVF conference on computer vision and pattern recognition},
  pages={4938--4947},
  year={2020}
}

@inproceedings{he2016deep,
  title={\href{https://openaccess.thecvf.com/content_cvpr_2016/papers/He_Deep_Residual_Learning_CVPR_2016_paper.pdf}{Deep residual learning for image recognition}},
  author={He, Kaiming and Zhang, Xiangyu and Ren, Shaoqing and Sun, Jian},
  booktitle={Proceedings of the IEEE conference on computer vision and pattern recognition},
  pages={770--778},
  year={2016}
}

@inproceedings{pang2022smog,
  title={\href{https://www.ecva.net/papers/eccv_2022/papers_ECCV/papers/136900264.pdf}{Unsupervised visual representation learning by synchronous momentum grouping}},
  author={Pang, Bo and Zhang, Yifan and Li, Yaoyi and Cai, Jia and Lu, Cewu},
  booktitle={European Conference on Computer Vision},
  pages={265--282},
  year={2022},
  organization={Springer}
}

@inproceedings{beeching2020learning,
  title={Learning to plan with uncertain topological maps},
  author={Beeching, Edward and Dibangoye, Jilles and Simonin, Olivier and Wolf, Christian},
  booktitle={European Conference on Computer Vision},
  pages={473--490},
  year={2020},
  organization={Springer}
}

@article{chen2020ask,
  title={\href{https://arxiv.org/pdf/2011.00517.pdf}{Ask your humans: Using human instructions to improve generalization in reinforcement learning}},
  author={Chen, Valerie and Gupta, Abhinav and Marino, Kenneth},
  journal={arXiv preprint arXiv:2011.00517},
  year={2020}
}

@inproceedings{li2020hrl4in,
  title={\href{http://proceedings.mlr.press/v100/li20a/li20a.pdf}{Hrl4in: Hierarchical reinforcement learning for interactive navigation with mobile manipulators}},
  author={Li, Chengshu and Xia, Fei and Martin-Martin, Roberto and Savarese, Silvio},
  booktitle={Conference on Robot Learning},
  pages={603--616},
  year={2020},
  organization={PMLR}
}

@inproceedings{vezhnevets2020options,
  title={\href{http://proceedings.mlr.press/v119/vezhnevets20a/vezhnevets20a.pdf}{Options as responses: Grounding behavioural hierarchies in multi-agent reinforcement learning}},
  author={Vezhnevets, Alexander and Wu, Yuhuai and Eckstein, Maria and Leblond, R{\'e}mi and Leibo, Joel Z},
  booktitle={International Conference on Machine Learning},
  pages={9733--9742},
  year={2020},
  organization={PMLR}
}

@inproceedings{le2018hierarchical,
  title={\href{http://proceedings.mlr.press/v80/le18a/le18a.pdf}{Hierarchical imitation and reinforcement learning}},
  author={Le, Hoang and Jiang, Nan and Agarwal, Alekh and Dud{\'\i}k, Miroslav and Yue, Yisong and Daum{\'e} III, Hal},
  booktitle={International conference on machine learning},
  pages={2917--2926},
  year={2018},
  organization={PMLR}
}

@inproceedings{vezhnevets2017feudal,
  title={\href{http://proceedings.mlr.press/v70/vezhnevets17a/vezhnevets17a.pdf}{Feudal networks for hierarchical reinforcement learning}},
  author={Vezhnevets, Alexander Sasha and Osindero, Simon and Schaul, Tom and Heess, Nicolas and Jaderberg, Max and Silver, David and Kavukcuoglu, Koray},
  booktitle={International Conference on Machine Learning},
  pages={3540--3549},
  year={2017},
  organization={PMLR}
}

@inproceedings{chane2021goal,
  title={\href{http://proceedings.mlr.press/v139/chane-sane21a/chane-sane21a.pdf}{Goal-conditioned reinforcement learning with imagined subgoals}},
  author={Chane-Sane, Elliot and Schmid, Cordelia and Laptev, Ivan},
  booktitle={International Conference on Machine Learning},
  pages={1430--1440},
  year={2021},
  organization={PMLR}
}

@article{wijmans2019dd,
  title={\href{https://arxiv.org/pdf/1911.00357.pdf}{Dd-ppo: Learning near-perfect pointgoal navigators from 2.5 billion frames}},
  author={Wijmans, Erik and Kadian, Abhishek and Morcos, Ari and Lee, Stefan and Essa, Irfan and Parikh, Devi and Savva, Manolis and Batra, Dhruv},
journal = {International Conference on Learning Representations},
  year={2019}
}

@article{chen2019behavioral,
  title={\href{https://roboticsproceedings.org/rss15/p10.pdf}{A behavioral approach to visual navigation with graph localization networks}},
  author={Chen, Kevin and De Vicente, Juan Pablo and Sepulveda, Gabriel and Xia, Fei and Soto, Alvaro and V{\'a}zquez, Marynel and Savarese, Silvio},
  journal={arXiv preprint arXiv:1903.00445},
  year={2019}
}

@inproceedings{he2020momentum,
  title={\href{https://openaccess.thecvf.com/content_CVPR_2020/papers/He_Momentum_Contrast_for_Unsupervised_Visual_Representation_Learning_CVPR_2020_paper.pdf}{Momentum contrast for unsupervised visual representation learning}},
  author={He, Kaiming and Fan, Haoqi and Wu, Yuxin and Xie, Saining and Girshick, Ross},
  booktitle={Proceedings of the IEEE/CVF conference on computer vision and pattern recognition},
  pages={9729--9738},
  year={2020}
}

@article{caron2020unsupervised,
  title={\href{https://proceedings.neurips.cc/paper_files/paper/2020/file/70feb62b69f16e0238f741fab228fec2-Paper.pdf}{Unsupervised learning of visual features by contrasting cluster assignments}},
  author={Caron, Mathilde and Misra, Ishan and Mairal, Julien and Goyal, Priya and Bojanowski, Piotr and Joulin, Armand},
  journal={Advances in neural information processing systems},
  volume={33},
  pages={9912--9924},
  year={2020}
}

@inproceedings{kwon2023renderable,
  title={\href{https://openaccess.thecvf.com/content/CVPR2023/papers/Kwon_Renderable_Neural_Radiance_Map_for_Visual_Navigation_CVPR_2023_paper.pdf}{Renderable Neural Radiance Map for Visual Navigation}},
  author={Kwon, Obin and Park, Jeongho and Oh, Songhwai},
  booktitle={Proceedings of the IEEE/CVF Conference on Computer Vision and Pattern Recognition},
  pages={9099--9108},
  year={2023}
}

@inproceedings{wasserman2023last,
  title={\href{https://proceedings.mlr.press/v205/wasserman23a/wasserman23a.pdf}{Last-mile embodied visual navigation}},
  author={Wasserman, Justin and Yadav, Karmesh and Chowdhary, Girish and Gupta, Abhinav and Jain, Unnat},
  booktitle={Conference on Robot Learning},
  pages={666--678},
  year={2023},
  organization={PMLR}
}

@inproceedings{habitat19iccv,
  title     =     {\href{https://openaccess.thecvf.com/content_ICCV_2019/papers/Savva_Habitat_A_Platform_for_Embodied_AI_Research_ICCV_2019_paper.pdf}{Habitat: {A} {P}latform for {E}mbodied {AI} {R}esearch}},
  author    =     {Manolis Savva and Abhishek Kadian and Oleksandr Maksymets and Yili Zhao and Erik Wijmans and Bhavana Jain and Julian Straub and Jia Liu and Vladlen Koltun and Jitendra Malik and Devi Parikh and Dhruv Batra},
  booktitle =     {Proceedings of the IEEE/CVF International Conference on Computer Vision (ICCV)},
  year      =     {2019}
}

@inproceedings{xiazamirhe2018gibsonenv,
  title={\href{https://openaccess.thecvf.com/content_cvpr_2018/papers/Xia_Gibson_Env_Real-World_CVPR_2018_paper.pdf}{Gibson env: real-world perception for embodied agents}},
  author={Xia, Fei and R. Zamir, Amir and He, Zhi-Yang and Sax, Alexander and Malik, Jitendra and Savarese, Silvio},
  booktitle={Computer Vision and Pattern Recognition (CVPR), 2018 IEEE Conference on},
  year={2018},
  organization={IEEE}
}

@inproceedings{zhu2017target,
  title={\href{https://ieeexplore.ieee.org/stamp/stamp.jsp?arnumber=7989381&casa_token=ILZrKW3zGGYAAAAA:ihJyQvWkjllptXNonkyPhUSU0ROV1edDqxGoHm74IH_1B59EBYbqTDClJFPpiDNUo2t4HmkyYiU}{Target-driven visual navigation in indoor scenes using deep reinforcement learning}},
  author={Zhu, Yuke and Mottaghi, Roozbeh and Kolve, Eric and Lim, Joseph J and Gupta, Abhinav and Fei-Fei, Li and Farhadi, Ali},
  booktitle={2017 IEEE international conference on robotics and automation (ICRA)},
  pages={3357--3364},
  year={2017},
  organization={IEEE}
}

@article{chaplot2020object,
  title={\href{https://proceedings.neurips.cc/paper/2020/file/2c75cf2681788adaca63aa95ae028b22-Paper.pdf}{Object goal navigation using goal-oriented semantic exploration}},
  author={Chaplot, Devendra Singh and Gandhi, Dhiraj Prakashchand and Gupta, Abhinav and Salakhutdinov, Russ R},
  journal={Advances in Neural Information Processing Systems},
  volume={33},
  pages={4247--4258},
  year={2020}
}

@inproceedings{mezghan2022memory,
  title={\href{https://ieeexplore.ieee.org/stamp/stamp.jsp?arnumber=9981090&casa_token=n04VmdV6gVsAAAAA:vHaLTPxRb1Bn3Lk7JqX9fHHc8SniEhpikjYHzVrnz2UilwTbIA79NMMORNuLmoo5WKhl-53z-kk}{Memory-augmented reinforcement learning for image-goal navigation}},
  author={Mezghan, Lina and Sukhbaatar, Sainbayar and Lavril, Thibaut and Maksymets, Oleksandr and Batra, Dhruv and Bojanowski, Piotr and Alahari, Karteek},
  booktitle={2022 IEEE/RSJ International Conference on Intelligent Robots and Systems (IROS)},
  pages={3316--3323},
  year={2022},
  organization={IEEE}
}

@inproceedings{al2022zero,
  title={\href{https://openaccess.thecvf.com/content/CVPR2022/papers/Al-Halah_Zero_Experience_Required_Plug__Play_Modular_Transfer_Learning_for_CVPR_2022_paper.pdf}{Zero experience required: Plug \& play modular transfer learning for semantic visual navigation}},
  author={Al-Halah, Ziad and Ramakrishnan, Santhosh Kumar and Grauman, Kristen},
  booktitle={Proceedings of the IEEE/CVF Conference on Computer Vision and Pattern Recognition},
  pages={17031--17041},
  year={2022}
}

@article{yadav2022offline,
  title={\href{https://arxiv.org/pdf/2204.13226.pdf}{Offline visual representation learning for embodied navigation}},
  author={Yadav, Karmesh and Ramrakhya, Ram and Majumdar, Arjun and Berges, Vincent-Pierre and Kuhar, Sachit and Batra, Dhruv and Baevski, Alexei and Maksymets, Oleksandr},
  journal={arXiv preprint arXiv:2204.13226},
  year={2022}
}

@inproceedings{fang2019scene,
  title={\href{https://openaccess.thecvf.com/content_CVPR_2019/papers/Fang_Scene_Memory_Transformer_for_Embodied_Agents_in_Long-Horizon_Tasks_CVPR_2019_paper.pdf}{Scene memory transformer for embodied agents in long-horizon tasks}},
  author={Fang, Kuan and Toshev, Alexander and Fei-Fei, Li and Savarese, Silvio},
  booktitle={Proceedings of the IEEE/CVF conference on computer vision and pattern recognition},
  pages={538--547},
  year={2019}
}

@inproceedings{beeching2020egomap,
  title={\href{https://arxiv.org/pdf/2002.02286.pdf}{EgoMap: Projective mapping and structured egocentric memory for Deep RL}},
  author={Beeching, Edward and Dibangoye, Jilles and Simonin, Olivier and Wolf, Christian},
  booktitle={Joint European Conference on Machine Learning and Knowledge Discovery in Databases},
  pages={525--540},
  year={2020},
  organization={Springer}
}

@article{kumar2018visual,
  title={\href{https://proceedings.neurips.cc/paper/2018/file/66368270ffd51418ec58bd793f2d9b1b-Paper.pdf}{Visual memory for robust path following}},
  author={Kumar, Ashish and Gupta, Saurabh and Fouhey, David and Levine, Sergey and Malik, Jitendra},
  journal={Advances in neural information processing systems},
  volume={31},
  year={2018}
}

@article{epstein2017cognitive,
  title={\href{https://www.nature.com/articles/nn.4656}{The cognitive map in humans: spatial navigation and beyond}},
  author={Epstein, Russell A and Patai, Eva Zita and Julian, Joshua B and Spiers, Hugo J},
  journal={Nature neuroscience},
  volume={20},
  number={11},
  pages={1504--1513},
  year={2017},
  publisher={Nature Publishing Group US New York}
}

@article{tolman1948cognitive,
  title={\href{http://www.guillaumegronier.com/2020-psychologiegenerale/resources/Tolman1948.pdf}{Cognitive maps in rats and men.}},
  author={Tolman, Edward C},
  journal={Psychological review},
  volume={55},
  number={4},
  pages={189},
  year={1948},
  publisher={American Psychological Association}
}

@article{peer2021structuring,
  title={\href{https://www.sciencedirect.com/science/article/abs/pii/S1364661320302503?casa_token=cLDDnkiULSkAAAAA:sK0IT3ofCoSmsJUK6I860jYJadtyl7O3g6eHCu0165nvuSKgHYo0lPc02Kbl54JK7yQymZnjWbA}{Structuring knowledge with cognitive maps and cognitive graphs}},
  author={Peer, Michael and Brunec, Iva K and Newcombe, Nora S and Epstein, Russell A},
  journal={Trends in cognitive sciences},
  volume={25},
  number={1},
  pages={37--54},
  year={2021},
  publisher={Elsevier}
}

@article{bar2024navigation,
  title={Navigation world models},
  author={Bar, Amir and Zhou, Gaoyue and Tran, Danny and Darrell, Trevor and LeCun, Yann},
  journal={arXiv preprint arXiv:2412.03572},
  year={2024}
}

@article{gervet2023navigating,
  title={\href{https://arxiv.org/pdf/2212.00922.pdf}{Navigating to objects in the real world}},
  author={Gervet, Theophile and Chintala, Soumith and Batra, Dhruv and Malik, Jitendra and Chaplot, Devendra Singh},
  journal={Science Robotics},
  volume={8},
  number={79},
  pages={},
  year={2023},
  publisher={American Association for the Advancement of Science}
}

@InProceedings{Gupta_2017_CVPR,
author = {Gupta, Saurabh and Davidson, James and Levine, Sergey and Sukthankar, Rahul and Malik, Jitendra},
title = {\href{https://openaccess.thecvf.com/content_cvpr_2017/papers/Gupta_Cognitive_Mapping_and_CVPR_2017_paper.pdf}{Cognitive Mapping and Planning for Visual Navigation}},
booktitle = {Proceedings of the IEEE Conference on Computer Vision and Pattern Recognition (CVPR)},
month = {July},
year = {2017}
}

@article{devo2020towards,
  title={\href{https://ieeexplore.ieee.org/document/9102361}{Towards generalization in target-driven visual navigation by using deep reinforcement learning}},
  author={Devo, Alessandro and Mezzetti, Giacomo and Costante, Gabriele and Fravolini, Mario L and Valigi, Paolo},
  journal={IEEE Transactions on Robotics},
  volume={36},
  number={5},
  pages={1546--1561},
  year={2020},
  publisher={IEEE}
}

@inproceedings{seymour2021maast,
  title={\href{https://ieeexplore.ieee.org/stamp/stamp.jsp?arnumber=9561058&casa_token=Zs45YmfCOv4AAAAA:RqGi5FO3k3Y9jrrxkdjGZOvyX2STk3NQWWzykT7lrFsMFoUm0szFPEKQVjEcErz1RpTQxDa50Qs&tag=1}{Maast: Map attention with semantic transformers for efficient visual navigation}},
  author={Seymour, Zachary and Thopalli, Kowshik and Mithun, Niluthpol and Chiu, Han-Pang and Samarasekera, Supun and Kumar, Rakesh},
  booktitle={2021 IEEE International Conference on Robotics and Automation (ICRA)},
  pages={13223--13230},
  year={2021},
  organization={IEEE}
}

@article{dayan1992feudal,
  title={\href{https://proceedings.neurips.cc/paper/1992/file/d14220ee66aeec73c49038385428ec4c-Paper.pdf}{Feudal reinforcement learning}},
  author={Dayan, Peter and Hinton, Geoffrey E},
  journal={Advances in neural information processing systems},
  volume={5},
  year={1992}
}

@article{mirowski2018learning,
  title={\href{https://proceedings.neurips.cc/paper_files/paper/2018/file/e034fb6b66aacc1d48f445ddfb08da98-Paper.pdf}{Learning to navigate in cities without a map}},
  author={Mirowski, Piotr and Grimes, Matt and Malinowski, Mateusz and Hermann, Karl Moritz and Anderson, Keith and Teplyashin, Denis and Simonyan, Karen and Zisserman, Andrew and Hadsell, Raia and others},
  journal={Advances in neural information processing systems},
  volume={31},
  year={2018}
}

@INPROCEEDINGS{Savarese-RSS-19, 
    AUTHOR    = {Kevin Chen AND Juan Pablo de Vicente AND Gabriel Sepulveda AND Fei Xia AND Alvaro Soto AND Marynel Vazquez AND Silvio Savarese}, 
    TITLE     = {\href{https://arxiv.org/pdf/1903.00445.pdf}{A Behavioral Approach to Visual Navigation with Graph Localization Networks}}, 
    BOOKTITLE = {Proceedings of Robotics: Science and Systems}, 
    YEAR      = {2019}, 
    ADDRESS   = {FreiburgimBreisgau, Germany}, 
    MONTH     = {June}, 
    DOI       = {10.15607/RSS.2019.XV.010} 
}

@article{bono2023learning,
  title={Learning with a Mole: Transferable latent spatial representations for navigation without reconstruction},
  author={Bono, Guillaume and Antsfeld, Leonid and Sadek, Assem and Monaci, Gianluca and Wolf, Christian},
  journal={arXiv preprint arXiv:2306.03857},
  year={2023}
}

@article{bono2023end,
  title={End-to-end (instance)-image goal navigation through correspondence as an emergent phenomenon},
  author={Bono, Guillaume and Antsfeld, Leonid and Chidlovskii, Boris and Weinzaepfel, Philippe and Wolf, Christian},
  journal={arXiv preprint arXiv:2309.16634},
  year={2023}
}

@article{ramakrishnan2022environment,
  title={Environment predictive coding for visual navigation},
  author={Ramakrishnan, Santhosh Kumar and Nagarajan, Tushar},
  journal={ICLR 2022},
  year={2022}
}

@inproceedings{li2024memonav,
  title={MemoNav: Working memory model for visual navigation},
  author={Li, Hongxin and Wang, Zeyu and Yang, Xu and Yang, Yuran and Mei, Shuqi and Zhang, Zhaoxiang},
  booktitle={Proceedings of the IEEE/CVF Conference on Computer Vision and Pattern Recognition},
  pages={17913--17922},
  year={2024}
}
}


\end{document}